\documentclass[letterpaper, 10 pt, conference]{ieeeconf}  

\IEEEoverridecommandlockouts                              

\usepackage{graphics} 
\usepackage{epsfig} 
\usepackage{mathptmx} 
\usepackage{times} 
\usepackage{amsmath} 
\usepackage{amssymb}  
\usepackage{cite}
\usepackage{algorithm}
\usepackage[usenames,dvipsnames]{color}
\usepackage{algpseudocode}
\usepackage{lipsum}
\usepackage{array}
\usepackage{multirow}

\title{\LARGE \bf
 Efficient Game-Theoretic Planning with Prediction Heuristic for Socially-Compliant Autonomous Driving
}

\author{Chenran Li$^{1}$, Tu Trinh$^{1}$, Letian Wang$^{2}$, Changliu Liu$^{2}$, Masayoshi Tomizuka$^{1}$ and Wei Zhan$^{1}$
\thanks{$^{1}$ Department of Mechanical Engineering, University of California, Berkeley, California, USA {\tt\small chenran\_li, tutrinh, tomizuka, wzhan@berkeley.edu}}%
\thanks{$^{2}$ Robotics Institute, Carnegie Mellon University, Pittsburgh, PA, USA {\tt\small letianw, cliu6@andrew.cmu.edu}}%
}

\begin{document}

\maketitle
\thispagestyle{empty}
\pagestyle{empty}

\begin{abstract}

Planning under social interactions with other agents is an essential problem for autonomous driving. As the actions of the autonomous vehicle in the interactions affect and are also affected by other agents, autonomous vehicles need to efficiently infer the reaction of the other agents. Most existing approaches formulate the problem as a generalized Nash equilibrium problem solved by optimization-based methods. However, they demand too much computational resource and easily fall into the local minimum due to the non-convexity. Monte Carlo Tree Search (MCTS) successfully tackles such issues in game-theoretic problems. However, as the interaction game tree grows exponentially, the general MCTS still requires a huge amount of iterations to reach the optima. In this paper, we introduce an efficient game-theoretic trajectory planning algorithm based on general MCTS by incorporating a prediction algorithm as a heuristic. On top of it, a social-compliant reward and a Bayesian inference algorithm are designed to generate diverse driving behaviors and identify the other driver's driving preference. Results demonstrate the effectiveness of the proposed framework with datasets containing naturalistic driving behavior in highly interactive scenarios.

\end{abstract}

\section{INTRODUCTION}
While rapid progress has been achieved on autonomous driving in recent years, tackling social interactions among autonomous vehicles and human drivers in real world remains one of the greatest challenges. 
Under social interactions, vehicles' mutual impact on others has to be captured. Otherwise, it may lead to either excessively conservative and inefficient interactions or catastrophic accidents due to insufficient social interaction understanding.

Game-theoretic planning frameworks \cite{lavalle2000robot,wang2021game,dreves2018generalized,li2017game} have been widely utilized to tackle social interactions, where the influence among agents is modeled in the system. These frameworks 
consider the interaction system including a robot vehicle called the ego agent and other interactive vehicles. Usually, the actions of other agents are formulated as a function of the ego agent's action, forming a closed-loop interaction system. To solve these game-theoretic interaction problems, many of the existing works \cite{Sadigh-RSS-16,sadigh2016information,fisac2019hierarchical} exploit optimization-based approaches. However, they are too computationally inefficient and would easily fall into a local minimum, especially for non-convex problems. 

To address such problems, Monte Carlo Tree Search (MCTS) has been considered in driving interaction problems \cite{sun2020game,kurzer2018decentralized}. Without the severe impact of the domain knowledge such as behavior assumptions or end conditions, MCTS can effectively evaluate the whole search space and successfully find the global optima.
However, as a typical sampling method, MCTS suffers from notorious inefficiency problems. As the tree expands, though the whole space is covered and the optimal solution can be searched on a fine-grained scale, a huge amount of memory and iterations are required. 

\begin{figure}[t!]
\centerline{\includegraphics[width=0.85\linewidth]{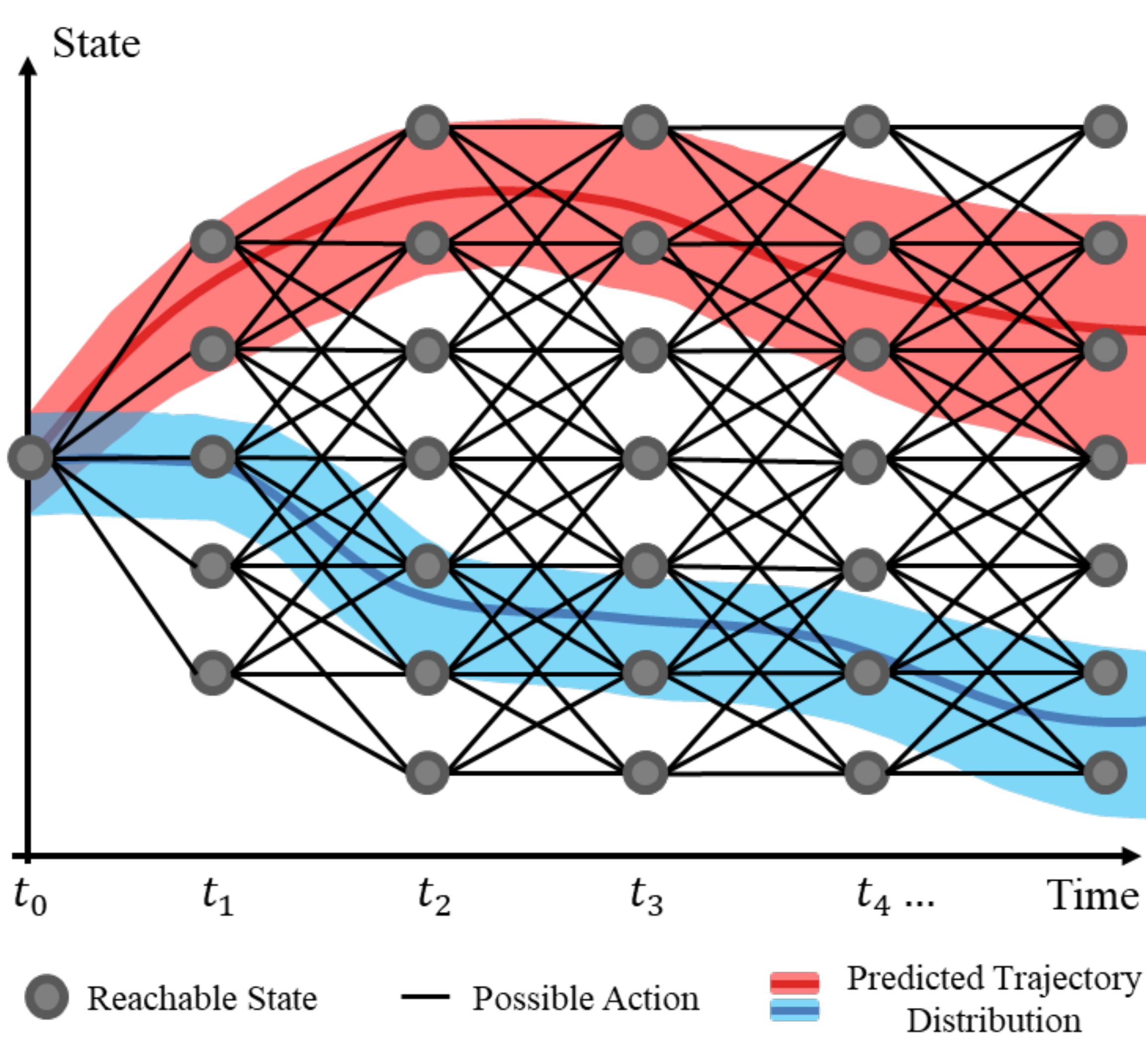}}
\caption{
Each node on the plot represents a state that can be visited by the searching algorithm, while the blue and red areas represent the suggested valuable area from the prediction models.}
\label{fig:insight}
\vspace{-1.5em}
\end{figure}

Nevertheless, our insight is that, as shown in Fig. \ref{fig:insight}, there are only a few branches that are truly valuable for searching in terms of dynamic feasibility, driving comfort/safety, interaction pattern, etc., though the tree grows rapidly. Such prior knowledge regarding which valuable areas should be searched can come from multiple sources. 

We resort to the easily accessed and well established behavior prediction models\cite{wang2022transferable,salzmann2020trajectron++,wang2021hierarchical,tripicchio2022modeling}, which can efficiently predict possible future behavior patterns. These behavior pattern predictions, though sometimes coarse or not highly accurate, can serve as an important heuristic to guide the search of MCTS, so that computational resources can be efficiently allocated at the valuable branches. However, the natural integration of the prediction heuristics into the search process is not trivial and requires delicate design. For example, overly strong prior knowledge would deprive MCTS of the capability to reach the globally optimal solution. Therefore, in this paper, we propose a novel prediction-heuristic tree search architecture to achieve efficient game-theoretic planning in interactive driving scenarios.

Moreover, in real-world interactions, human drivers may exhibit diverse social attributes. For example, in terms of driving style, while some drivers are courteous, there are also aggressive drivers \cite{schwarting2019social,wang2021socially,sun2018courteous}. To operate safely and efficiently in these social interactions, the autonomous vehicles should be able to: 
1) exhibit behaviors with diverse driving styles to adapt to different situations and explicitly express its own driving intentions to others; and
2) identify others' driving styles online for conflict resolution and collision avoidance at an early stage. 
However, such social behavior has not been captured in previous MCTS-based driving methods, in which all the agents are modeled with the same fixed policy. Such modeling is lacking in the ability to generate diverse driving styles and recognize others' driving styles, resulting in severe safety or efficiency issues in intensive social interactions. In this paper, we design a socially-compliant reward capturing the selfish and courteous driving preference, by which driving behavior with diverse styles could be generated. A Bayesian inference algorithm is utilized to identify other drivers' driving styles online by estimating the parameter in the socially-compliant reward.

The key contributions of this work are as follows.
\begin{itemize}
    \item Proposing a new framework by incorporating prediction heuristic into MCTS architecture for efficient game-theoretic planning in interaction-aware driving scenarios.
    \item Designing a social-compliant reward and a Bayesian inference algorithm within the MCTS architecture, to generate diverse driving behaviors and identify others' driving styles online.
    \item Conducting comprehensive experiments with naturalistic data, which demonstrates the effectiveness of prediction heuristic and the game-theoretic interaction in highly interactive driving scenarios.  
\end{itemize}

To the best of our knowledge, this paper is the first to incorporate prediction heuristic and socially-compliant reward into the MCTS framework for socially interactive driving tasks.

\section{Problem Formulation}
\label{sec:problem formulation}

In this work, we consider the interaction system with two agents including a robot vehicle called the ego agent and an interactive vehicle called the opponent agent. The superscript $E$ denotes the ego agent and $O$ denotes the opponent agent. The interaction system satisfies an overall dynamics: 
\begin{equation}
\begin{aligned}
    \textbf{x}_{t+1} = f \left(\textbf{x}_{t},u_{t}^{E},u_{t}^{O}\right ), 
\end{aligned}
\label{overall dynamics}
\end{equation}
where $\textbf{x}_t =  \{\textbf{x}^{E}_t,\textbf{x}^{O}_t\}$ is the joint state of the ego agent and the opponent agent at time step $t$ representing their dynamic states such as position, velocity and etc. $u_t$ is the control input at time step $t$ representing the action of each agent such as acceleration, steering angle and etc.
Let $N$ be the horizon of the interaction. The action sequence of the each vehicle in the next $N$ steps are denoted as $U_{N}^{E/O} = [u_{0}^{E/O},...,u_{N-1}^{E/O}]^T$. 

The agents in the interaction are optimizers that their control inputs maximize the cumulative reward function of each own noted by $R^{E}$ and $R^{O}$. Thus, the optimal action sequence $U_N^*$ that each vehicle will take can be represented as:
\begin{equation}
\begin{aligned}
    & {U_N^{E}}^* = \arg \max_{U_N^{E}} R^{E}\left(\textbf{x}_0, U_N^{E},{U_N^{O}}^*\right), \\
    & {U_N^{O}}^* = \arg \max_{U_N^{O}} R^{O}\left(\textbf{x}_0,{U_N^{E}}^*,U_N^{O}\right).
\end{aligned}
\label{eq:Ego and opponent optimization}
\end{equation}
\begin{figure}[t!]
\centerline{\includegraphics[width=0.85\linewidth]{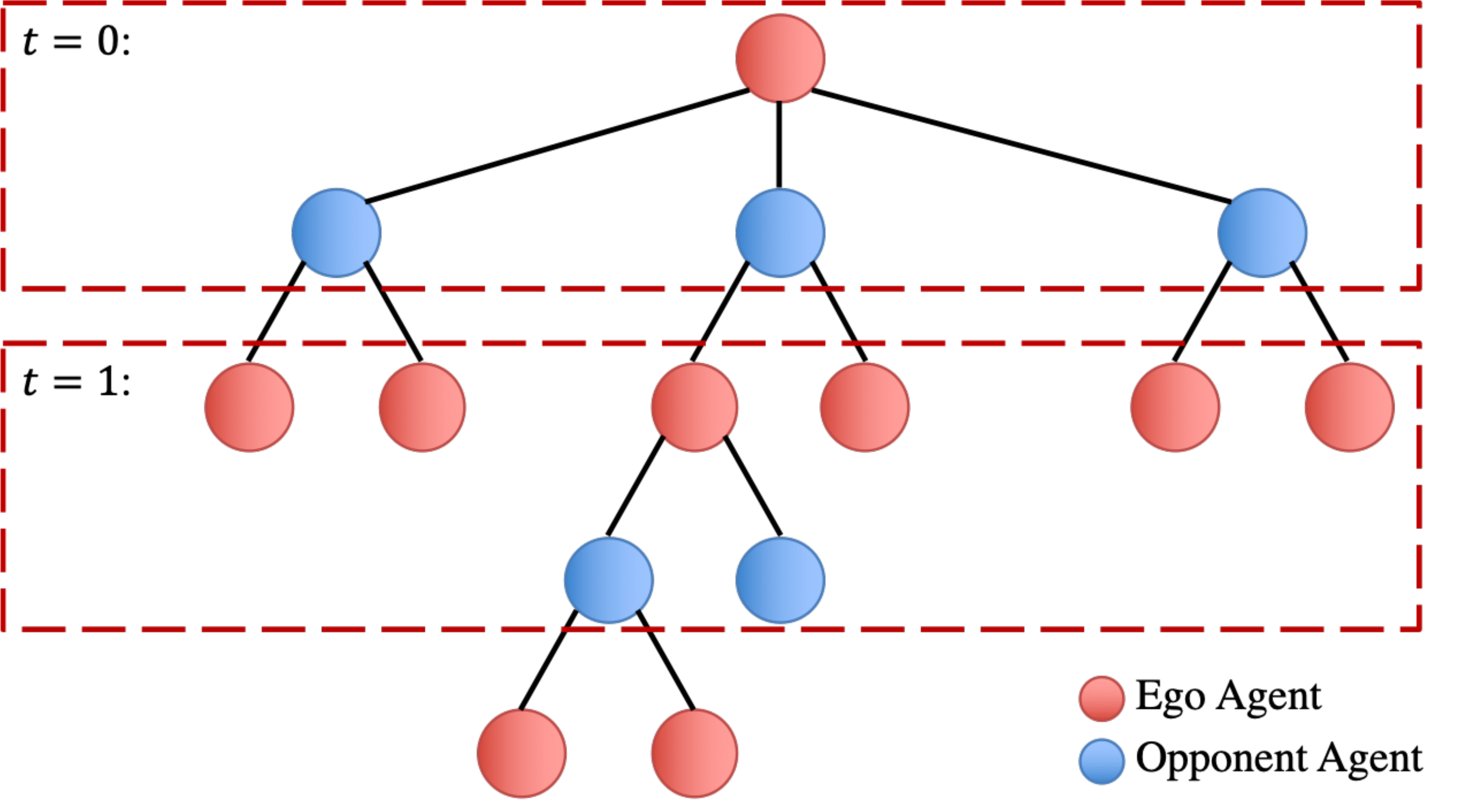}}
\caption{Game Tree Representation of Interaction}
\label{fig:GameTree}
\vspace{-1.5em}
\end{figure}
To better explain the diverse driving preferences of the opponent agent and generate socially-compliant behavior for the ego agent, the courtesy factor should be considered in the reward function during the planning. Similar to \cite{Sadigh-RSS-16,levine2012continuous}, the reward for modeling the agents in the proposed method is designed as a linear combination of the egoism reward term and the courtesy reward term. For instance, the ego agent's reward is formulated as:
\begin{equation}
\begin{aligned}
    \label{eq:social reward}
    R^{E}\left(\textbf{x}_0, U_N^{E},{U_N^{O}}\right) &=  \gamma \, R^{E}_\text{Egoism}\left(\textbf{x}_0, U_N^{E},{U_N^{O}}\right) \\
    & + (1-\gamma) \, R^{E}_\text{Courtesy}\left(\textbf{x}_0, U_N^{E},{U_N^{O}}\right),
\end{aligned}
\end{equation}
where $R^{E}_\text{Egoism}$ and $R^{E}_\text{Courtesy}$ are the egoism reward term and the courtesy reward term respectively, a detailed definition of which can be found in Sec~\ref{sec:reward design}. $\gamma \in [0,1]$ balances the reward features.

As in \cite{wang2021socially,sun2018courteous,sadigh2016information}, we further assume that two agents are running a Stackelberg game where one agent is the leader and the other agent is a follower. Note that the Stackelberg game assumption is reasonable for most interaction scenarios since different agents always have different right-of-way in interaction. Thus we can always select the vehicle with a higher right-of-way as the leader. Without losing generality, in this paper we assume the ego agent as the leader agent. 

Therefore, the optimization in Eq.~\ref{eq:Ego and opponent optimization} can be reformed as:
\begin{equation}
\begin{aligned}
    & {U_N^{E}}^* = \arg \max_{U_N^{E}} R^{E}\left(\textbf{x}_0, U_N^{E},{U_N^{O}}^*(U_N^{E})\right), \\
    & {U_N^{O}}^*(U_N^{E}) = \arg \max_{U_N^{O}} R^{O}\left(\textbf{x}_0,{U_N^{E}},U_N^{O}\right).
\end{aligned}
\label{Ego and opponent optimization}
\end{equation}

To solve the optimization problem, we regard the optimal action sequence as the optimal node path in the game tree shown in Fig. \ref{fig:GameTree}, where the two agents make decisions sequentially at each search step. Formally, with the discretized action space, at search iteration $t$, we have an ego node $n_{t}^{E}=(\textbf{x}_t,U_t^{E},U_t^{O})$ and an opponent node $n_{t}^{O}=(\textbf{x}_t,U_{t+1}^{E},U_t^{O})$. The root node and the terminal node can be represented as $(\textbf{x}_0,\emptyset,\emptyset)$ and $(\textbf{x}_N,U_N^{E},U_N^{O})$ respectively.

\begin{table}[t]
\caption{Notations in Algorithm \ref{alg:PredictionHeuristic}}
\label{tab:Variable table}
\setlength{\arrayrulewidth}{0.25mm}
\centering
  \begin{tabular}{c c}
   \hline
   Notation & Description \\
    \hline
    $\textbf{x}_0$ & Current state of the interaction system  \\
    $\hat {\textbf{Y}}_{N}^i$ & $i^{th}$ prediction trajectory with $N$ steps  \\
    $\hat{ \textbf{y}}^i_t$ & Predicted state in $\hat {\textbf{Y}}_{N}^i$ at step $t$ \\
    $G_t^i$ & Confident range introduced by $\hat{ \textbf{y}}^i_t$ \\
    $G$ & Set of all confident ranges \\
    $w_\text{conf}$ & Confident weight\\
    $U^{E/O}_N(n)$ & Action sequence stored in terminal node\\
    $I_{unsafe}(n)$ & Indicator function of safety constraints\\
    $Q^{E/O}(n)$ & Total reward of all roll-out trajectories\\
    $Q^{E/O}_s(n)$ & Total searching reward of all roll-out trajectories\\
    $q^{E/O}_s$ & Reward of a roll-out trajectory\\
    $q^{E/O}$ & Searching reward of a roll-out trajectory\\
    \hline
  \end{tabular}
\end{table}

When searching for optimal action sequences in such game tree problems, Monte Carlo Tree Search (MCTS) has achieved significant success in various applications, especially in multi-agent setting\cite{best2020decentralised,kartal2016monte,browne2012survey}. However, the search space grows exponentially as the action space and the time space extend, which leads to low efficiency and infeasible practical deployment due to computation limitations. Therefore, in this paper, we propose the prediction-heuristic search architecture based on general MCTS for efficient searching and planning.

\section{Prediction Heuristic Search Architecture}

\begin{figure}[t!]
\centerline{\includegraphics[width=0.75\linewidth]{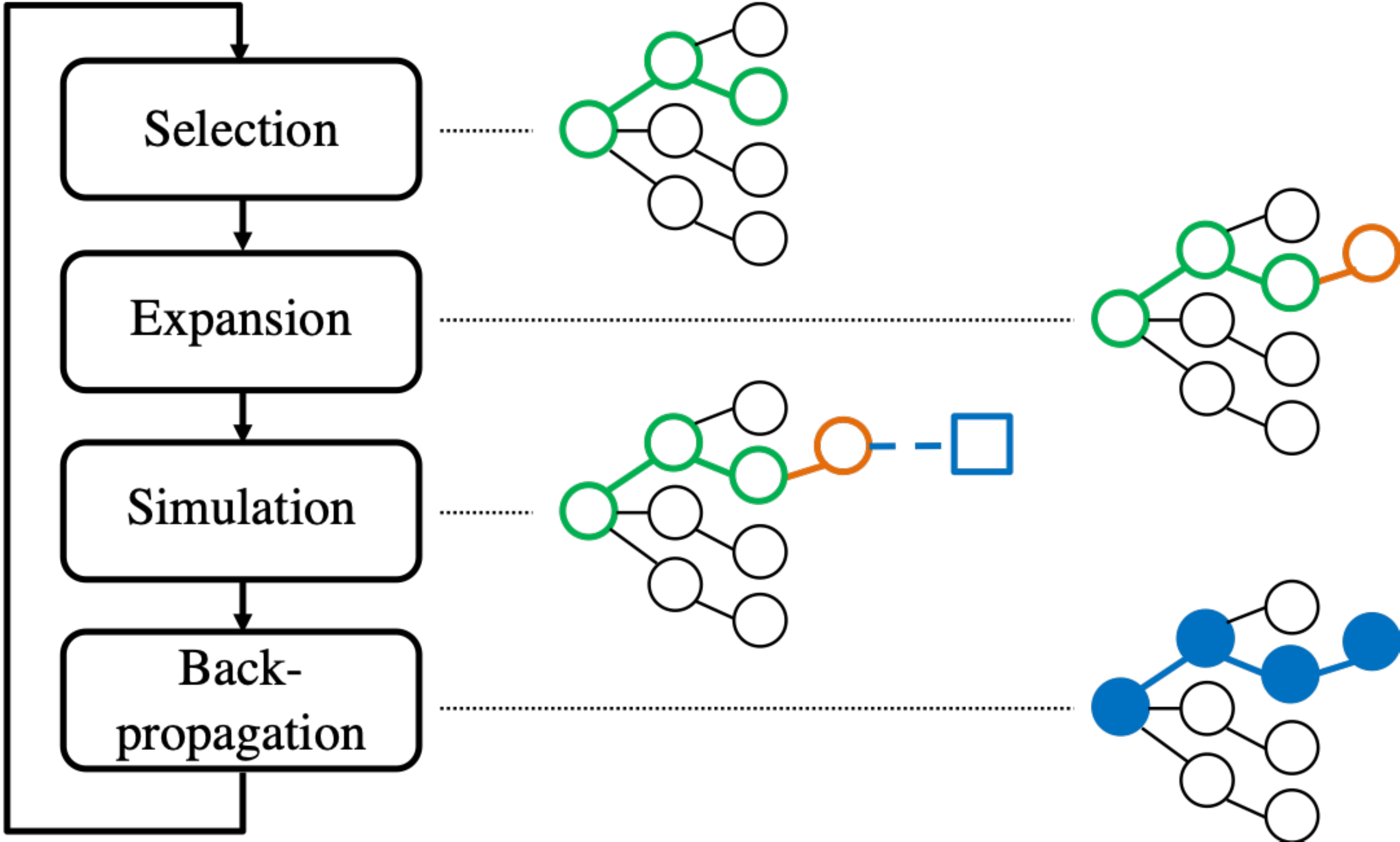}}
\caption{The Structure of The General MCTS Algorithm}
\label{MCTS}
\end{figure}

\begin{algorithm}[t]
    \caption{Prediction Heuristic Search Architecture}
    \label{alg:PredictionHeuristic}
    \textbf{Input:}
	Current state $\textbf{x}_0$,  K predicted trajectories $\hat {\textbf{Y}}_{N}^i=\{\hat{ \textbf{y}}^i_1,...,\hat {\textbf{y}}^i_N\}, i=1,...,K$. \newline
	\textbf{Output:} Root node $n_0^E$ of the built tree
	\begin{algorithmic}[1]
	\State Initialize $G = \{G^1_1,...G^K_N\}$ from $\hat {\textbf{Y}}_{N}^i$, for $i=1,...,K$ and $t=1,...,N$.
	\State Initialize the root node $n_0^E = (\textbf{x}_0,\emptyset,\emptyset)$.
	\State Initialize iteration counter  $Iter = 0$.
    \While {$Iter < \text{Maximum Iterations}$}
        \State $n_l \leftarrow \Call{Selection}{n_0^E}$
        \If {$n_l$ is non-terminal}
            \State $n_N,U_N^{E},U_N^{O} \leftarrow \text{Roll-out}(n_l)$
        \Else
            \State $n_N,U_N^{E},U_N^{O} \leftarrow n_l, U_N^{E}(n_l), U_N^{O}(n_l)$
        \EndIf 
        \If {$I_{unsafe}(n_N)$}
            \State $q^E, q^O \leftarrow 0,0$
        \Else
            \State $q^E, q^O \leftarrow R^{E}(\textbf{x}_0, U_N^{E},{U_N^{O}}), R^{O}(\textbf{x}_0, U_N^{E},{U_N^{O}})$
        \EndIf
        \While {$n_l$ is not NULL} 
            \State $C(n_l)\leftarrow C(n_l) + 1$
            \State $w_\text{conf} \leftarrow 1$
            \If {$n_l$ is a ego node}
                \State $w_\text{conf} \leftarrow w_\text{conf}(n_l,G)$
            \EndIf
            \State $q_s^E, q_s^O \leftarrow  w_\text{conf} \times q^E, w_\text{conf} \times q^O$
            \State $Q^E(n_l), Q^E_s(n_l) \leftarrow Q^E(n_l) + q^E, Q^E_s(n_l) +q_s^E$
            \State $Q^O(n_l), Q^O_s(n_l) \leftarrow Q^O(n_l) + q^O, Q^O_s(n_l) +q_s^O$
            \State $n_l \leftarrow \text{parent of } n_l $
        \EndWhile
    \EndWhile
    \State \Return $n_0^E$
    \end{algorithmic}
\end{algorithm}
\setlength{\textfloatsep}{0.1cm}
\setlength{\floatsep}{0.1cm}
\begin{algorithm}[t]
    \caption{Function for Algorithm~\ref{alg:PredictionHeuristic}}
    \label{alg:Functions}
    \begin{algorithmic}[1]
    \Function{Selection}{$n_t$}
        \While{$n_t$ is non-terminal} 
            \If {$n_t$ is not fully expanded}
                \State $n_l \leftarrow $ expand $n_t$ with an untried action.
                \While{$I_{unsafe}(n_l)$}
                    \State $n_l \leftarrow $ expand $n_t$ with an untried action.
                \EndWhile
                \State \Return $n_l$
            \Else
                   \State $n_l \leftarrow$ solving Eq.~\ref{eq:selection} with $Q_s^E$ when $n_t$ is ego node; $Q_s^O$ when $n_t$ is opponent node.
            \EndIf
        \EndWhile
    \EndFunction
    \end{algorithmic}
\end{algorithm}
\setlength{\textfloatsep}{0.1cm}
\setlength{\floatsep}{0.1cm}

As shown in Fig. \ref{MCTS}, the general MCTS algorithm includes mainly four steps per search iteration that are selection, expansion, simulation, and back-propagation\cite{chaslot2008monte}. At the selection phase, the algorithm will recursively select the most valuable child node until reaching the leaf node. After selecting the node, the algorithm will expand the tree at the selected node according to the available actions. During the simulation phase, the algorithm will use a roll-out policy to simulate the game from the expanded node to the end. At the back-propagated phase, the algorithm will extract the rewards from the roll-out sample and use them to update the average rewards through the selected nodes. After reaching the defined maximum iterations, the solution can be extracted by selecting the maximum rewards child node for each layer.

As the MCTS gradually evaluates the whole interaction action space, the game tree grows exponentially, which would result in either failure in finding optimal solutions or computationally-infeasible in practical deployment. 
However, in terms of dynamics feasibility, driving comfort/safety, and interaction pattern, there are in fact only a few branches that are truly meaningful to search. Thus, our idea here is to use the predicted future trajectory, which is easily accessed from well-developed prediction models, to guide the exploration toward fruitful branches of the tree. The pseudo-code of the proposed method is presented in Algorithm \ref{alg:PredictionHeuristic}, \ref{alg:Functions}, where the variables used in the algorithm are summarized in the Table \ref{tab:Variable table} and will be defined in detail in the following subsections.

Similar to the general MCTS, let $C(n_t)$ denote the number of times that the input node has been visited. $Q^E(n_t), Q^O(n_t)$ denotes the total reward of all roll-out trajectories that pass the node $n_t$ for the ego agent and the opponent agent respectively. To keep the reward unbiased when extracting the planned trajectory, we introduce the searching rewards. Let $Q^E_s(n_t), Q^O_s(n_t)$ denote the total searching reward of all roll-out trajectory that pass the node $n_t$ for ego agent and opponent agent respectively. The selection policy is defined as 
\begin{equation}
\label{eq:selection}
\begin{aligned}
     \arg\max & {Q^{E/O}_s(n_l) \over C(n_l)} + c\sqrt{{2\ln{C(n_t)}}\over{C(n_l)}}.\\
     s.t.~& n_l \in \text{children of } n_t
\end{aligned}
\end{equation}
 The first term in the equation is the average reward of passing node $n_l$ to encourage the exploitation of high-reward actions. The second term is an exploration bonus to encourage the exploration of less-visited actions, where $c$ is a constant parameter to balance the trade-off between exploitation and exploration\cite{browne2012survey}.  

\subsection{Prediction Heuristic Exploration}
Within the MCTS process, searching is conducted for the ego agent and opponent agent iteratively as in Fig. \ref{fig:GameTree}. The searching nodes for the opponent agent are actually embedding a prediction module inside the planner to simulate the whole interaction. Though precise and explainable solutions can be searched, the search space is evaluated uniformly. Due to the fact that the possible interaction and driving patterns of the two vehicles are usually limited, the computation resources are mostly spared in uninformed search space, leading to low efficiency. On the other hand, nowadays many efficient learning-based prediction methods have been developed\cite{wang2021hierarchical,salzmann2020trajectron++,wang2022transferable}. 
With actual driver interaction captured in real human data, the predicted trajectory from these learned prediction models usually presents feasible interaction patterns of the two agents, like one yielding car and one passing car. Though predicted trajectories may not be as precise and accurate as that in the tree, they are still highly valuable prior knowledge that provides a good heuristic for the search. To integrate such heuristics into the search process while keeping sufficient exploration, we design a confidence range around the prediction trajectory and the associated confidence weight to guide exploration.

\subsubsection{Confidence range}
From the prediction model, we can retrieve $K$ possible future trajectories and the corresponding probabilities for the opponent agent, which may represent one interaction or driving pattern. Let $\hat {\textbf{Y}}_{N}^i=\{\hat{ \textbf{y}}^i_1,...,\hat {\textbf{y}}^i_N\}$ and $p_i$ denote the $i^{th}$ predicted trajectory and its probability of being executed respectively. Therefore, the state at time step $t$ can be modeled as a random distribution with mean $\hat {\textbf{y}}^i_t$ and variances $\Sigma^i_t$, where the variances $\Sigma^i_t$ in the predicted trajectory can be calculated by methods discussed in \cite{sun2021complementing, lakshminarayanan2016simple}. 
Thus, each predicted trajectory introduces a confidence range $G_t^i$ at each time step by Mahalanobis distance, which is given as 
\begin{equation}
\begin{aligned}
    G_t^i = \{ \textbf{x}_t^{O} | (\textbf{x}_t^{O} - \hat {\textbf{y}}^i_t )^T{\Sigma^i_t}^{-1}(\textbf{x}_t^{O} - \hat {\textbf{y}}^i_t) \leq \rho \},
\end{aligned}
\end{equation}
where $\rho$ is a constant parameter that controls the size of $G_t^i$. Note that after choosing parameter $\rho$, the size of $G_k^i$ only depends on the variance of the prediction trajectory. When the predicted trajectory has a large variance which means large uncertainty exists in the prediction heuristic, the size of $G_k^i$ will also increase, encouraging exploration in a larger space.

\subsubsection{Confidence weight}
With the confidence range, we introduce the confidence weight to balance the trade-off between maximizing reward and following the prediction. When back-propagating the reward to each layer, the searching reward of the roll-out sample $q_s^{E/O}$ will be weighted by its confidence weight $w_\text{conf}$ as in line 18 in Algorithm 1. To encourage wider exploration, the confidence weight should remain uniform in the whole confidence range instead of focusing on the mean. Branches inside multiple confidence ranges should also raise more explore attention. Based on these insights, $w_\text{conf}$ is then defined as the summation of associated probabilities $p_i$ of all confidence ranges that node $n_t$ is inside, namely
\begin{equation}
\begin{aligned}
    & w_\text{conf}(n_t,G) = \sum_{i \in V} p_i, \\
    & V = \{i|\textbf{x}_t^{O}(n_t) \in  G_t^i, G_t^i \in G \},
\end{aligned}
\end{equation}
where $G$ is the set of all confidence ranges.
By multiplying the reward with the confidence weight, the algorithm will explore more those branches that are more interaction-feasible in practice and have a higher reward. \vspace{-0.05em}
\subsection{Roll-out Policy}
In the simulation phase, the algorithm will use a roll-out policy to simulate the game from the expanded node to the end. In this process, efficiently generating valuable sample trajectories is important for the effective evaluation of the expanded nodes. Our insight is that the prediction heuristic introduced earlier in the expansion phase can also serve as valuable prior knowledge in this simulation phase. Therefore, we also exploit the prediction heuristic in the roll-out policy. 
Specifically, when generating a roll-out trajectory from an expanded node,  the algorithm will check the confidence range and related prediction trajectory that the node belongs to. Then the roll-out trajectory will be generated within confidence ranges introduced by the same prediction trajectory, by a random action selection process. 
Meanwhile, in terms of driving comfort, the roll-out action is constrained by comfort jerk $j_\text{comf}$, namely   
\begin{equation}
\begin{aligned}
    \left| a_{t+1} - a_{t} \right| \leq  j_\text{comf},
\end{aligned}
\end{equation}
where $a_t$ is the corresponding acceleration of the action $u_t$.

\subsection{Safety Constraints}
During searching, identifying unsafe nodes is not only important for the feasibility of the result but also for increasing the searching efficiency. Let $I_{unsafe}(n_t)$ be the indicator function of safety constraints that will be true when the trajectory between the root node and input node violates the given safety constraints such as collision with other agents and road edge. There are two procedures in which the safety constraints will be evaluated.
\subsubsection{Sample evaluation}
As shown at line 11 in Algorithm 1, the safety constraints will be evaluated before calculating the reward of the roll-out sample. If the sample trajectory is unsafe, the reward of it will be fixed to zero, which penalizes the average reward of the branch for safety but save the branch for remaining potential trajectories.
\subsubsection{Expanding tree}
After expanding nodes, it is necessary to check if they satisfy the safety constraints, as at line 5 in Algorithm 2. If the indicator function gives that the expanded node is unsafe, all the succeeding branches after it will be blocked by this unsafe node. Therefore, the algorithm will remove the unsafe node when expanding the tree to reduce the searching space. 

\section{Socially-Compatible Driving}
Safe and efficient interaction necessitates the inclusion of diverse social preferences in driving. Thus in this section, we elaborate on the socially-compatible reward design and the utilized Bayesian inference algorithm to online estimate the reward parameter.
\begin{algorithm}[t]
	\caption{Reward Parameter Bayesian Inference}
	\label{alg:policy inference}
	\textbf{Input:}
	M reward candidate parameter samples ${\gamma^i}$,  corresponding weights ${\omega}^i{_{k-r-1}}$, and observations $\hat{\textbf{x}}{_{k-r:k}}$\newline
	\textbf{Output:}
	Updated weights ${\omega}{_{k-r}}$, estimated parameter $\gamma_{k-r}$.
	\begin{algorithmic}[1]
		\If{k-r-1=0}
    		\State Initialize candidate reward parameter samples ${\gamma^i}$ and corresponding weights $\omega_{k-r-1}$ \label{code:initial}
		\EndIf
		\For {all M candidate reward parameter samples}
    		\State Update ${\omega}{_{k-r}^{i}} \leftarrow {\omega}{_{k-r-1}^{i}} \times
    		p(\widehat{\textbf{x}}{_{k-r:k}\vert\gamma}{^i})$, Eq. (\ref{eq:update probability})
    		\label{code:update}
		\EndFor
		\State Normalize ${{\omega}}{_{k-r} \leftarrow {{\omega}}{_{k-r}/\sum\nolimits^M_{i=1}}{\omega}{_{k-r}^{i}}}$
		\State Compute ${\gamma}_{k-r} \leftarrow\sum\nolimits^M_{i=1}\gamma{^{i}{\omega}{_{k-r}^{i}}}$
		\label{code:estimate}
	\end{algorithmic}
\end{algorithm}
\subsection{Socially-compatible reward design}
\label{sec:reward design}
As introduced in Eq.~\ref{eq:social reward}, for one selected ego vehicle, the socially-compatible reward is designed as the linear combination of egoism reward term $R^{E}_\text{Egoism}(\textbf{x}_0, U_N^{E},{U_N^{O}})$ and the courtesy reward term $R^{E}_\text{Courtesy}(\textbf{x}_0, U_N^{E},{U_N^{O}})$. The parameter $\gamma$ balances the two terms, which can be utilized to generate behaviors with different styles as shown in Sec~\ref{sec:exp.B}. The egoism reward is defined to capture an agent's own utilities:
\begin{equation}
\label{eq:cost function}
    R_{Egoism}^E({\bf{x}}_0, U_N^E, U_N^O) =  \theta^T\sum_{t=0}^{N-1}\phi({\bf{x}}_t, U_t^E, U_t^O),
\end{equation}
where $\phi{\in}\mathrm{R}^3$ denotes the three egoism utility feature: efficiency, comfort, and safety. $\theta{\in}\mathrm{R}^3$ denotes their relative weights, which can be manually tuned or learned via Inverse Reinforcement Learning\cite{levine2012continuous,ziebart2008maximum}. 
Inspired by \cite{schwarting2019social}, the courtesy reward term can be represented by the opponent agent's egoism reward, that is
\begin{equation}
\begin{aligned}
    R^{E}_\text{Courtesy}\left(\textbf{x}_0, U_N^{E},{U_N^{O}}\right)=  R^{O}_\text{Egoism}\left(\textbf{x}_0, U_N^{E},{U_N^{O}}\right).\\
\end{aligned}
\end{equation}
 
\subsection{Reward Parameter Estimation}
Human behaviors are naturally heterogeneous, time-varying, and stochastic \cite{abuduweili2021robust,wang2021online}. Driving preferences may vary in individuals and circumstances. Hence, we should infer the opponent agent's current driving policy by estimating the reward parameter $\gamma$ in Eq.~\ref{eq:social reward}.

The insight of reward update is that, though human drivers could not communicate directly, their historic behaviors could leak their mental state, which can be leveraged to infer their driving preferences. Technically, based on maximum entropy, the probability of one candidate reward parameter $\gamma$ is proportional to the probability of the observed trajectory $\hat{\bf{x}}_{k-r:k}$ conditional on that candidate reward parameter:

\begin{equation}
\label{eq:bayesian}
p({\bf{\gamma}}|{\hat{\bf{x}}}_{k-r:k}) \propto p({\hat{\bf{x}}}_{k-r:k}|{\bf{\gamma}})p({\bf{\gamma}}),
\end{equation}
where we look $r$ steps to the past from current time step $k$, and $p({\bf{\gamma}})$ is the prior probability. The probability of the observed trajectory under one candidate reward parameter is proportional to its reward under that candidate reward parameter:
\begin{align}
\label{eq:update probability}
p({\hat{\bf{x}}}_{k-r:k}|\gamma)\approx \frac{e^{R({\hat{{U}}_{k-r:k}}, \gamma)}}{\sum_{{{U}\in \textbf{U}}}{e^{R({{{U}}_{k-r:k}}, \gamma)}}},
\end{align}
where $\textbf{U}$ is the set of all possible actions, and $\hat{{U}}$ is the action resulting in the observed trajectory. $R({{{U}}_{k-r:k}}, \gamma)$ denotes the reward gained under action ${{{U}}_{k-r:k}}$ and parameter $\gamma$, as detailed in Eq.~\ref{eq:social reward}.

With Eq.~\ref{eq:bayesian} and Eq.~\ref{eq:update probability}, we formulate a Bayesian inference algorithm to estimate the reward parameter for one agent on the fly. 
The algorithm is outlined in Algorithm \ref{alg:policy inference}. Line~\ref{code:initial} samples M reward parameters $\gamma^i$ and initializes corresponding weights $\omega^i(\gamma^i)$. Line~\ref{code:update} updates weights for each sample. Line~\ref{code:estimate} estimates reward parameter $\gamma_{k-r}$ as the mean of the posterior distribution, which is taken as the reward parameter in the current time step $\gamma_k$ by the assumption of a small change of reward parameter $\gamma$ over the observed horizon $N$.

\begin{table*}[ht]
\caption{Visiting Times for Different Depth in Search Tree}
\label{tab:Depth table}
\centering
  \begin{tabular}{|c|c|c|c|c|c|c|c|c|c|c|}
    \hline
    \multicolumn{2}{|c|}{Depth} & Layer 1 & Layer 2 & Layer 3 & Layer 4 & Layer 5 & Layer 6 & Layer 7 & Layer 8 & Layer 9  \\
    \hline
    \multirow{2}{7.6em}{General~MCTS} 
    & $V_\text{Max}$ & 5077 & 853 & 142 & 24 & 4 & 1 & 0 & 0 & 0 \\
    & $V_\text{Other}$  & 4984 & 844 & 142 & 23 & 4 & 0 & 0 & 0 & 0  \\
    \hline
    \multirow{2}{7.6em}{Proposed~Method} 
    & $V_\text{Max}$ & 5259 & 3036 & 547 & 464 & 87 & 49 & 8 & 2 & 1 \\
    & $V_\text{Other}$  & 4948 & 444 & 497 & 16 & 75 & 7 & 8 & 1 & 0  \\
    \hline
    \multicolumn{2}{|c|}{Ratio of $V_\text{Max}$ Between Methods}  & 1.04 &  3.55 & 3.85 & 19.33 & 21.75 & 49.00 & Inf. & Inf. & Inf.  \\
    \hline
  \end{tabular}
\end{table*}

\begin{figure}[t]
\centerline{\includegraphics[width=0.9\linewidth]{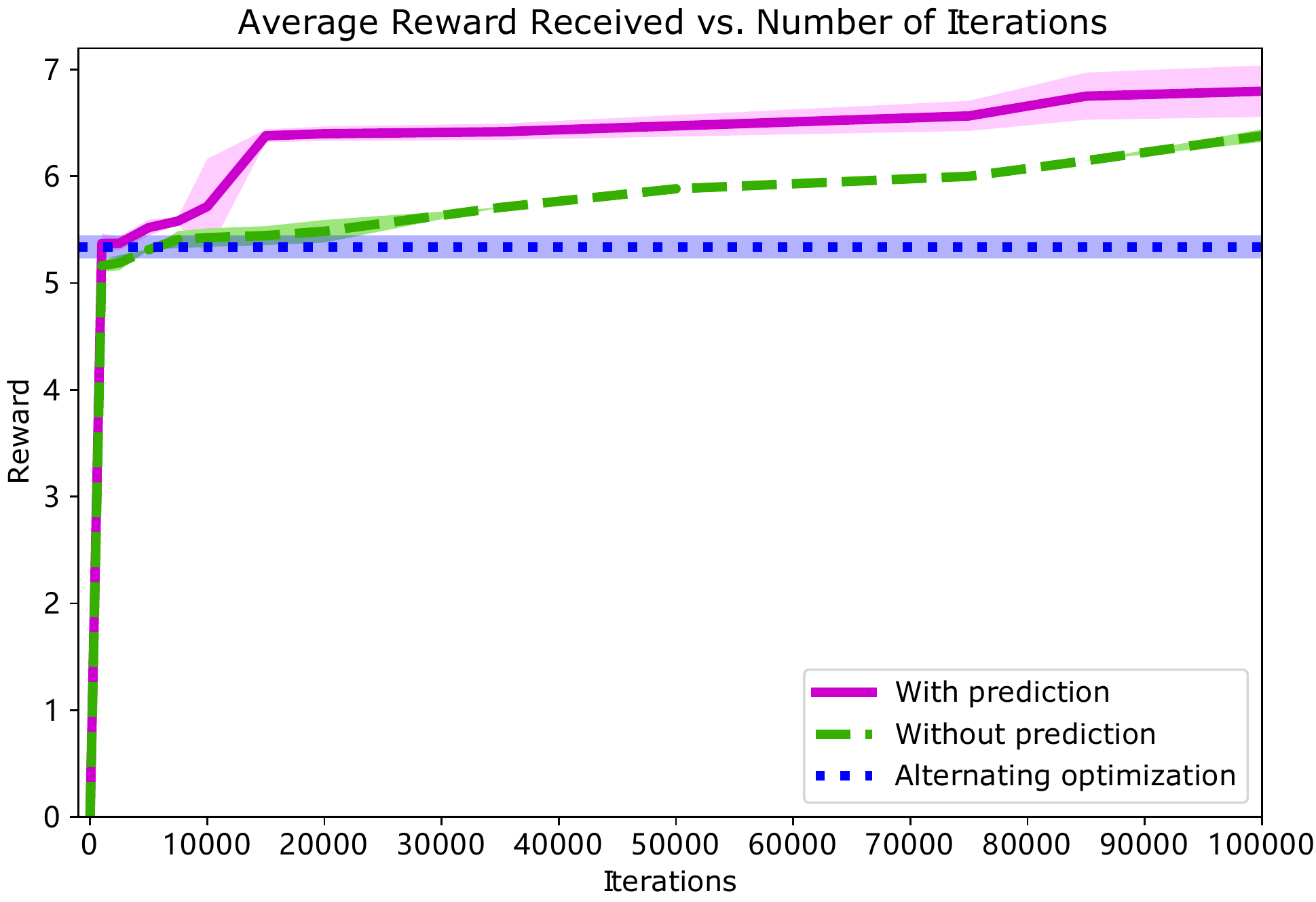}}
\caption{Average Reward of The Output Trajectory with Different Maximum Iterations}
\label{fig:rewards}
\end{figure}

\begin{figure*}[t]
\centerline{\includegraphics[width=0.90\linewidth]{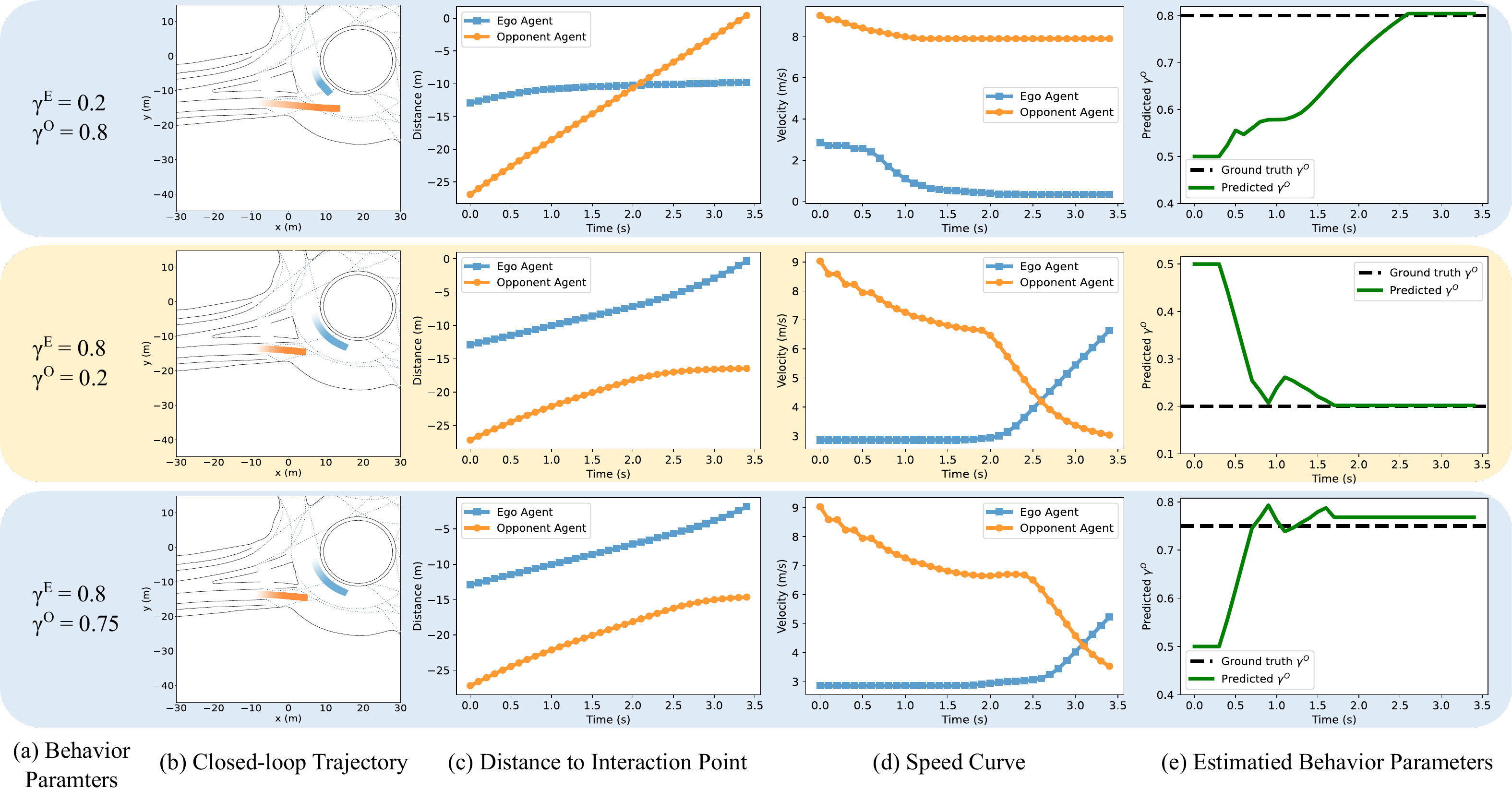}}
\caption{Closed-loop Simulation with Different Driving Behaviors. In (b), cars at earlier time steps are drawn more transparently for time-lapse display.}
\label{fig:cl}
\vspace{-1.5em}
\end{figure*}

\section{Experiments}
In this section, a series of experiments were conducted to test the performance of the proposed algorithm. In Sec~\ref{sec:exp.A}, we analyze the optimality and efficiency improvement brought by the proposed algorithm in real human interaction scenarios. The detailed searching information is provided showing the effectiveness. In Sec~\ref{sec:exp.B}, we deploy the planner in several closed-loop simulations with different behaviors generated by the designed socially-compatible reward. 
In Sec~\ref{sec:exp.C}, we further discuss the implementation usage of the proposed framework.

All experiments were generated from real-world driving scenarios from the INTERACTION Dataset \cite{zhan2019interaction}.
There are different kinds of scenarios in the INTERACTION dataset, and we utilized the roundabout scenario for our experiments. The insight behind such a choice is that interaction essentially happens between agents that are driving toward the same region and thus have a conflict with each other\cite{markkula2020defining}. In such a general definition, interaction can be modeled regardless of scenarios. Consequently, our studies and conclusions can also apply to other driving scenarios, like intersections, merging scenarios, etc.

Let $l_\text{ref}$ denote the length of reference path, $v_\text{max}$ denote the maximum legal speed. The states of an agent were represented with the longitudinal position $s_t \in [0,\,l_\text{ref}]$  and longitudinal speed $v_t  \in [0,\,v_\text{max}]$ along the reference path. The agent's control input was represented as longitudinal acceleration $a_t$ that was discretized into a list with choices as $[-3, -2, -1, 0, 1, 2]\text{m}/\text{s}^2$. The planning time step was $0.5$s and the planning horizon $N = 5$. The reward function in these experiments was given by
\begin{equation}
\label{eq:cost function in experiment}
    R_{Egoism} = \theta^T \sum_{t=0}^{N-1} \left[e^{- \alpha a_t^2},  1 - e^{- \beta v_{t+1}^2}  \right]^T ,
\end{equation}
where $\alpha$, $\beta$ are constant parameters. 
The experiments were mainly simulating around the interaction point which is defined as the cross point of the reference paths of the ego agent and opponent agent.

\subsection{Searching Performance}
\label{sec:exp.A}
In this subsection, we randomly selected a time frame before any agent reach the interaction point from the dataset, and planned the future trajectory for ten times with different random seeds by different algorithms.  To focus our study on the effect of the prediction heuristic, the predicted trajectory was generated by adding the ground truth trajectory with Gaussian noise ${N}(0,0.4^2)$ meter for each trajectory point.

\subsubsection{Baselines}
To evaluate the effectiveness of the proposed algorithm, we first implemented a baseline MCTS algorithm without the prediction heuristic, which is used in \cite{sun2020game}. Additionally, we also implemented the alternative optimization algorithm based on the iterative best response optimization algorithm\cite{schwarting2019social}. The alternating optimization algorithm will initialize each agent with a random action sequence and iteratively search for the best action sequence of each agent separately until both agents have converged. 

\subsubsection{Performance Comparison}
As shown in Fig. \ref{fig:rewards}, we first present the reward of the planned trajectory generated by the three different methods, as the search iteration increases. Note that the alternating optimization algorithm iteratively searches in two trees until both trees converge (one tree for one agent), and the other two methods only search in one tree where both two agents are nested. Consequently, one iteration in the alternating optimization algorithm represents different computation costs compared to the other two methods. For a fair comparison, we only draw the reward of the final trajectory generated by the alternating optimization algorithm.
As in Fig. \ref{fig:rewards}, the two methods that search within one tree result in higher rewards than the alternating optimization algorithm. Also, benefit from the prediction heuristic, our proposed method shown with the purple curve converges faster than the general MCTS algorithm shown with the green curve, demonstrating the efficiency improvement of our method.

\subsubsection{Searching analysis} To further investigate the effect of prediction heuristic in detial, we also compared the visiting times of the maximum reward node $V_\text{Max}$ and other nodes $V_\text{Other}$ in the same layer and maximum depth that the algorithm was able to get. Table~\ref{tab:Depth table} presents the result with $30000$ iterations. Following the formulation in the previous sections, the odd layer contained nodes reached by the ego agent's actions while the even layer contained nodes reached by the opponent agent's actions.

As shown in the Table~\ref{tab:Depth table}, both the general MCTS and the proposed algorithm visited more times on the optimal node they find. As the searching layer gets deeper, the distribution of visiting nodes becomes more uniform for the general MCTS. However, with the prediction heuristic, the proposed algorithm was able to keep the focus on the optimal branches, especially in the layer of the opponent agent's actions. Moreover, as a benefit, the proposed algorithm was also able to search deeper and provide a longer plan than the general MCTS. Meanwhile, as it maintained more samples on the optimal node, it was more convincing on the optimality of the found trajectory. 

\subsection{Closed-Loop Simulation}
\label{sec:exp.B}
We conducted several closed-loop simulations to investigate the behaviors with different sets of the reward and evaluate the effectiveness of the reward parameter estimation. To generate a reactive game, both the ego agent and the opponent were planned by the proposed algorithm. The opponent agent was assumed to have knowledge of the ego vehicle's behavior parameter. The ego vehicle, however, had to figure out the opponent agent's setting with the proposed reward estimation. The simulation time step was $0.1$ s. To focus our study on the proposed method, the predicted trajectory was generated by the ground truth with Gaussian noise $N(0,0.4^2)$ meter at each trajectory point. Fig. \ref{fig:cl} shows three simulations conducted with different reward settings.

\subsubsection{Behavior diversity}
In the first simulation, the ego agent was set to be a courteous driver while the opponent agent was set to be an egoistic driver. Even though the ego agent was the leader in the game, the algorithm chose to yield the opponent agent since it received more rewards from the courtesy term. 

We then switched the behavior parameter of the ego agent and opponent agent to create the second simulation. As expected, the opponent agent behaved courteously and decided to yield the ego agent even with a higher velocity. 

To further study the effect of the behavior parameter, in the third experiment, both agents were set as egoistic drivers, but the ego agent was more egoistic than the opponent. As shown in the speed curves, compared with the courteous case, at the time step of $2$ s, the opponent refused to decrease its speed initially. Meanwhile, for the ego agent, after realizing the opponent's behavior preference, it decided to insist on its right-of-way by increasing its speed for a little after the time step of $1.7$ s. After the opponent eventually decided to yield at the time step of $2.6$ s, the ego agent was able to drive with a higher speed in safe.

\subsubsection{Behavior parameter estimation}
As shown with the (e) column in Fig. \ref{fig:cl}, the algorithm was able to accurately estimate the behavior parameter of the opponent agent for all simulations. In the last two simulations, the estimation was able to reach the ground truth in $1$ s, namely, $6$ time steps after receiving enough observations. However, in the first simulation, the convergence speed was relatively slow. The reason for it is mainly from the ego agent's behavior. As the ego agent chose to yield and stopped at an early stage, there was not much interaction that affect the opponent's actions, which made all $\omega^i$ update more slowly than in other cases. 

\section{Discussion}
\label{sec:exp.C}
In this section, we discuss the extension of the proposed algorithm for general scenarios and several implementation usages of the proposed algorithm in addition to decision and planning.
\subsection{Multi-agents Interactions}
The extension of the proposed formulation and solution to multi-agent settings can be defined in \textit{theory} without heavy effort. Following the same assumption of the Stackelberg game, the $M$ agents can be sorted by their right-of-way. Then, each interaction time step in the game tree shown in Fig. \ref{fig:GameTree} can be extended to include $M$ layers by the sorted order. Each layer represents the possible choices of actions for one agent at one time step. The proposed prediction heuristic can be applied similarly for each predicted agent. 

Though the extension takes little effort in the formulation, the multi-agent setting is still a challenging problem in practice, as the computational resources are required exponentially with the number of agents. 
Due to these limitations, the multi-agent problem is still hard to solve with issues of efficiency, convergence, and local optima. We believe this is an important and interesting future work that we will look into in future studies.

\subsection{Practical Usage of the Proposed Method}
\begin{figure}[t]
\centerline{\includegraphics[width=1.0\linewidth]{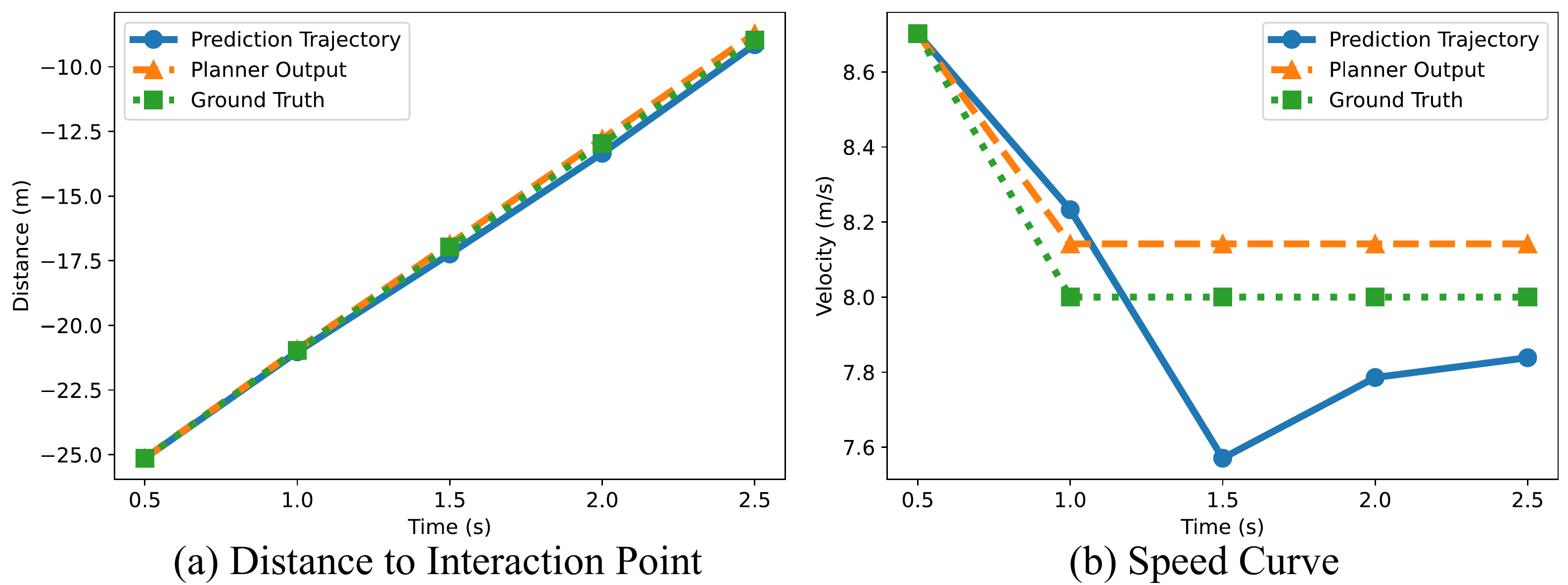}}
\caption{Comparison of the inferred trajectory and its corresponding prediction trajectory}
\label{fig:Comparison}
\end{figure}

During the closed-loop simulation, we also compared the inferred opponent's trajectory with its ground truth. As shown in Fig. \ref{fig:Comparison}, the inferred trajectory is able to recover the ground truth trajectory from its prediction with noises. Therefore, the proposed algorithm can also be used as an interaction-aware predictor to facilitate the learned model to generate better joint predictions.

\section{CONCLUSIONS}
In this paper, we proposed a game-theoretic planning algorithm by incorporating the prediction heuristic to make the search much more computationally efficient. A social-compliant reward and a Bayesian inference algorithm built on it were designed to generate diverse driving behaviors and identify others' driving preferences. As demonstrated in the results, the proposed planning framework was able to generate different behaviors by the designed reward and plan more efficiently than the general MCTS algorithm. The proposed Bayesian inference algorithm can successfully infer others' driving styles from the observations.




\section*{ACKNOWLEDGMENT}

The authors would like to thank Ran Tian for insightful advice and help.

\bibliographystyle{IEEEtran}
\bibliography{ref}{}

\end{document}